## Article

# The Use of a Large Language Model for Cyberbullying Detection

Bayode Ogunleye [1,*] and Babitha Dharmaraj [2]

1 Department of Computing & Mathematics, University of Brighton, Brighton BN2 4GJ, UK
2 Department of Digital, Data and Technology, Ofgem, London E14 4PU, UK
* Correspondence: b.ogunleye@brighton.ac.uk

**Abstract:** The dominance of social media has added to the channels of bullying for perpetrators. Unfortunately, cyberbullying (CB) is the most prevalent phenomenon in today's cyber world, and is a severe threat to the mental and physical health of citizens. This opens the need to develop a robust system to prevent bullying content from online forums, blogs, and social media platforms to manage the impact in our society. Several machine learning (ML) algorithms have been proposed for this purpose. However, their performances are not consistent due to high class imbalance and generalisation issues. In recent years, large language models (LLMs) like BERT and RoBERTa have achieved state-of-the-art (SOTA) results in several natural language processing (NLP) tasks. Unfortunately, the LLMs have not been applied extensively for CB detection. In our paper, we explored the use of these models for cyberbullying (CB) detection. We have prepared a new dataset (D2) from existing studies (Formspring and Twitter). Our experimental results for dataset D1 and D2 showed that RoBERTa outperformed other models.

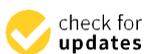

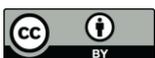

## 1. Introduction

The emergence of social technologies like Facebook, Twitter, TikTok, WhatsApp, Threads, and Instagram has improved communication amongst the people and businesses across the globe. However, despite the huge advantage of these platforms, they have also added channels of bullying for perpetrators. Cyberbullying (CB), often referred to as online bullying, is becoming an important issue that requires urgent attention. For illustration, in the USA, the Pew Research Center [1] reported that around two-thirds of US adolescents have been subjected to cyberbullying. Statista [2] reported in their survey that 41% of adults in the USA had experienced cyberbullying. The Pew Research Center [3] reported that 46% of teens in the USA aged 13 to 17 have been cyberbullied. The Office for National Statistics [4] reported that 19% of children aged 10 to 15 (this equates to 764,000 children) have experienced cyberbullying in England and Wales. Patchin and Hinduja [5] found out that 90% of tweens (9 to 12 years old) utilise social media or gaming apps, and 20% of tweens are involved in CB either as a victim, an offender, or a bystander.

This problem of cyberbullying (CB) is a relatively new trend that has recently gained more popularity as a subject. Cyberbullying is repetitive, aggressive, targeted, and intentional behaviour aimed at hurting an individual's or a group's feelings through an electronic medium [6,7]. CB takes many forms, including flaming, harassment, denigration, impersonation, exclusion, cyberstalking, grooming, outing, and trickery [6,8]. Cyberbullies are more likely to be technologically astute than physically stronger, making them better able to access victims online, conceal their digital footprints, and become involved in posting rumours, insults, sexual comments, threats, a victim's private information, or derogatory labels [9]. The fundamental causes of any bullying incident are the imbalance of power and the victim's perceived differences in race, sexual orientation, gender, socioeconomic level, physical appearance, and mannerism. Xu et al. [10] stated that CB participants could play the role of either a bully, victim, bystander, bully assistant, reinforcer, reporter, or accuser. Prior studies found out that CB impacts anxiety [11–13], depression [14,15], social isolation [16], suicidal thoughts [17–19], and self-harm [20,21]. Messias et al. [22] stated victims of cyberbullying have higher rates of depressive illnesses and suicidality than victims of traditional bullying. Patchin and Hinduja [5] stated CB victims admit that they frequently feel awkward or afraid to attend school, and it impacts their academic performance. In addition, they found out that nearly 70% of teens who reported being victims of cyberbullying stated it had a negative impact on their self-esteem, and nearly one-third claimed it had an impact on their friendships. Despite the impact and increasing rate of CB, unfortunately, there is limited attention paid to developing sophisticated approaches for automatic CB detection.

CB studies are yet to extensively explore the use of large language models (LLMs) for CB detection [23]. CB is commonly misinterpreted, leading to flawed systems with little practical use. Additionally, several studies only evaluated using swear words to filter CB, which is only one aspect of this topic, and swear words may not always indicate bullying on platforms with a high concentration of youngsters [6,24]. Thus, it is practically useful for developers and media handlers to have a robust system that understand context better, to enhance CB detection. In our study, we aim to evaluate the performance of large language models for CB detection. Unfortunately, there are some obstacles to CB detection. One is the issue of unavailable balanced and enriched benchmark datasets [6,23,25]. The issue of class imbalance has been a popular problem in machine learning (ML) applications, as the ML algorithms tend to be biased towards the majority class [26]. Past studies emphasised on the class imbalance problem in the CB context [27]. In most studies, the proportion of bullying posts is in the range of 4–20% of the entire dataset compared to non-bullying posts [6,28–30]. This opens the need to create a new, enriched dataset with balanced classes for effective CB detection and make it publicly available. To this end, we propose the use of Robustly optimized BERT approach (RoBERTa), a pre-trained large language model for cyberbullying detection. Thus, our contributions can be summarised as follows. We prepared a new dataset (D2) from existing studies for the development of algorithms on CB detection. We conducted an experimental comparison of sophisticated machine learning algorithms with two datasets (D1 and D2). We ascertained RoBERTa as the state-of-the-art (SOTA) method for automated cyberbullying detection. The rest of the paper is organised as follows. Section 2 will review the literature to provide background knowledge to this study. Section 3 will present the methodology. Section 4 will present and discuss the results, and Section 5 will provide conclusions and recommendations.

## 2. Related Work

Cyberbullying (CB) is the most prevalent phenomenon in today's digital world, and is a severe threat to the mental and physical health of cybercitizens [14,31]. Several studies have proposed various techniques for automated CB detection. For example, the authors in [10] crawled 1762 tweets from Twitter using keywords such as "bully, bullied, bullying". The data was labelled by five human annotators such that 684 were labelled as bullying and 1078 as non-bullying. They compared four traditional machine learning models, namely Naïve Bayes (NB), Support Vector Machines (linear SVM), Support Vector Machines (RBF—SVM), and Logistic regression (LR). Their result showed linear SVM achieved the best performance with a 77% F1 score. Agrawal and Awekar [28] compared machine learning (ML) models, namely Naïve Bayes (NB), Support Vector Machines (SVM), random forest (RF), convolutional neural network (CNN), long short-term memory (LSTM), bidirectional long short-term memory (BiLSTM), and BiLSTM with an attention mechanism. They used datasets from three different social media platforms namely, Formspring (a total of 12,773, split into 776 bully and 11,997 non-bullying text), which was collected by the authors in [8], Twitter (16,090, split into 1937 bullying through racism, 3117 bullying through sexism, and 11,036 non-bullying text), collected by the authors in [32], and Wikipedia (115,864, split into 13,590 attack and 102,274 non-attack text), collected by authors in [33]. They oversam-pled the minority class using the Synthetic Minority Oversampling technique (SMOTE). Their BiLSTM with attention implementation achieved an F1 score of at least 87% for the bullying across the three different social media platforms. Similarly, the authors in [34] reproduced the experiment in Agrawal and Awekar [28] with the Formspring dataset [8] only. Their results showed that SVM performed better than logistic regression (LR), decision tree (DT), and random forest (RF), with 98% accuracy and an F1 score of 93% (86% F1 score for bullying class). Alduailaj and Belghith [35] compared SVM and NB for cyberbullying detection in an Arabic language context. They collected 30,000 Arabic comments on 5 February 2021 from Twitter and YouTube, and the comments were labelled manually as bullying and non-bullying using most common and frequent Arabic bullying keywords detected from the comments. Their result showed term frequency inverse document frequency (TF-IDF) vectors deployed to SVM achieved accuracy of 95% and an F1 score of 88%. Lepe-Faúndez et al. [36] proposed a hybrid approach for CB (aggressive text) detection in a Spanish language context. They compared twenty-two hybrid models, a combination of lexicons and machine learning algorithms, across three datasets, namely Chilean (1013 aggressive and 1457 non-aggressive tweets), Mexican (2112 aggressive and 5220 non-aggressive tweets), and Chilean–Mexican (3127 aggressive and 6675 non-aggressive tweets) corpora. In their experiment, they tested the approaches with 30% of the dataset and showed a hybrid approach with lexicon features achieved superior performance, and models with SVM as classifier also achieved better performance amongst the ML algorithms deployed. Dewani et al. [37] showed that SVM and embedded hybrid N-gram approach performed best in detecting cyberbullying in the Roman Urdu language context, with an accuracy of 83%. Suhas-Bharadwaj et al. [38] applied extreme learning machine to classify cyberbullying messages, and achieved accuracy of 99% and an F1 score of 91%. Woo et al. [39] conducted a systematic review of CB literature. Their literature review findings suggest SVM and Naïve Bayes (NB) are the best-performing models for CB detection.

Recently, there has been development of large language models (LLMs), which have taken the world by surprise. The LLMs have been applied to several NLP tasks like topic modelling [40], sentiment analysis [41], a recommendation system [42], and harmful news detection [43]. In the context of CB detection, Paul and Saha [44] compared bidirectional encoder representations from transformers (BERT) to BiLSTM, SVM, LR, CNN, and a hybrid of RNN and LSTM, using three real-life CB datasets. The datasets are from Formspring [8], Twitter [32], and Wikipedia [33]. They used SMOTE to rebalance the dataset and, thus, showed BERT outperformed other models across the datasets with an F1 score of at least 91%. Similarly, Yadav et al. [45] applied BERT to the Formspring dataset [8] and Wikipedia dataset [33]. Their approach achieved an F1 score of 81% for the Wikipedia dataset. They rebalanced the Formspring dataset thrice, and achieved an F1 score of 59% with the first oversampling rate, an F1 score of 86% with the second oversampling dataset, and a 94% F1 score in the third oversampling dataset. However, it is worth mentioning that they have tested their model on the oversampled dataset, and thus might not be reliable in terms of generalisation. Yi and Zubiaga [41] used the same dataset from Formspring [8], Wikipedia [33], and Twitter [32]. They proposed XP-CB, a novel cross-platform adversarial framework based on transformers and adversarial learning models for cross-platform CB detection. They showed that XP-CB could enhance a transformer leveraging unlabelled data from the source and target platforms to come up with a common representation while preventing platform-specific training. They showed XP-CB achieved an average macro F1 score of 69%. In summary, popular data sources for CB detection are Wikipedia, Formspring, and Twitter [8,27,44,45]. Our literature review findings suggest that very few studies have used the transformers models (pre-trained large language models) for CB detection. A CB literature survey conducted by Woo et al. [39] found that most studies have used traditional machine learning models for CB detection. Thus, our paper looks to compare the performance of SOTA language models for CB detection.

## 3. Methodology

We propose the use of the fine-tuned Robustly optimized BERT approach (RoBERTa) for automatic cyberbullying (CB) detection. We conducted an experimental comparison of large language models to traditional machine learning models, such as support vector machine (SVM) and random forest (RF).

### 3.1. Bidirectional Encoder Representations from Transformers (BERT)

BERT [46] is a self-supervised autoencoder (AE) language model developed by Google in 2018 for training NLP systems. BERT is a bidirectional transformer-based model pre-trained on a large-scale Wikipedia corpus using the Masked Language Model (MLM) and next-sentence prediction tasks. BERT (base) is implemented based on transformer and attention mechanism, which uses encoders to read in input and decoders to output. The BERT base model consists of 12 layers of transformer blocks, a hidden layer size of 768, and 12 self-attention heads. The model comprises of two stages, which are the pre-training stage and the fine-tuning stage. Without making significant task-specific architecture alterations, the pre-trained BERT model can be improved with just one extra output layer to produce cutting-edge models for a variety of tasks, including question answering and language inference. The objective of the masked language model is to predict the actual vocabulary identity of a masked word based only on its context after randomly masking some of the tokens from the input. The MLM's intent permits the representation to combine the left and the right context, in contrast to the left-to-right language model pre-training, which allows us to pre-train a deep bidirectional Transformer. To fine-tune the pre-trained BERT model, the model is first instantiated with default parameters (used when pre-trained), and then the parameters are fine-tuned using labelled data from downstream tasks (text classification in our case). Every sequence will start with a particular classification token as the first token ([CLS]). For classification tasks, the last hidden state matching to this token is used as the aggregate sequence representation (Figure 1 below). The sum of the token embeddings, segmentation embeddings, and position embeddings constitutes the input embeddings, as shown in Figure 2 below.

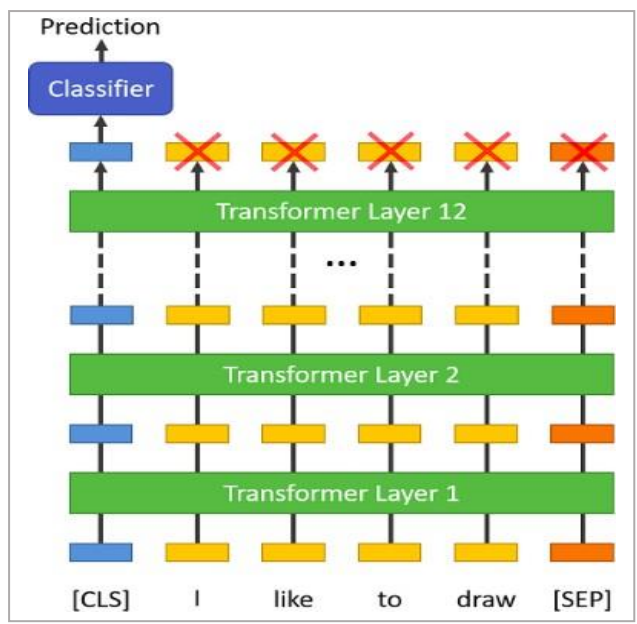

**Figure 1.** An illustration of BERT model architecture.

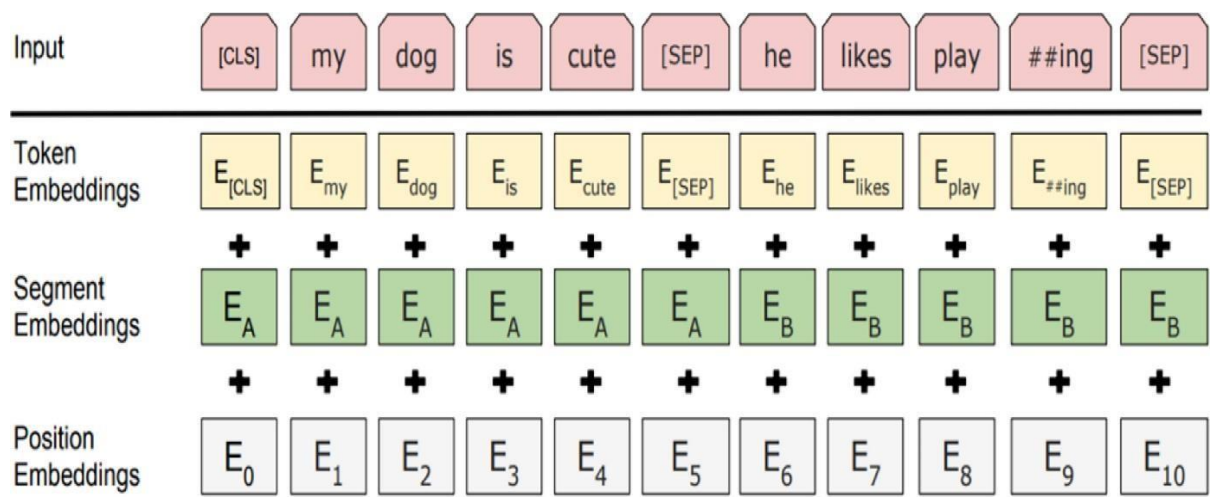

**Figure 2.** BERT input representation [46].

*3.2. RoBERTa*

Facebook AI Research (FAIR) identified the limitations of Google's BERT and proposed the Robustly optimized BERT approach (RoBERTa) in 2019. Liu et al. [47] stated that BERT was undertrained, and they modified the training method by (i) using a dynamic masking pattern instead of static, (ii) training with more data with large batches, (iii) removing next sentence prediction, and (iv) training on longer sentences, and proposed RoBERTa. As a result, RoBERTa outperforms BERT in terms of the masked language modelling objective and performs better on downstream tasks. For training the model, the researchers employed pre-existing unannotated NLP datasets, as well as the unique collection CC-News, which was compiled from publicly available news stories. RoBERTa is a component of the effort by Facebook to improve the state-of-the-art in self-supervised models so that they may be created with less dependency on time and data-labelling.

*3.3. XLNet*

XLNet [48] is a permutation based autoregressive transformer that combines the finest aspects of autoencoding and autoregressive language modelling while seeking to get around their drawbacks. BERT (as an autoencoder language model) ignores the dependency between the masked positions and leads to a pretrain–finetune discrepancy because it relies on masking the input to corrupt it. On the other hand, conventional autoregressive (AR) language models predict the next word based on the word's context, either in forward or backward direction, but not in both. XLNet's training objective calculates the likelihood of a word based on all possible word permutations in a sentence, rather than only those to the left or right of the target token. Integrating Transformer-XL and a carefully thought-out two-stream attention mechanism are just two of the ways that the XLNet neural architecture is designed to perform in perfect harmony with the autoregressive (AR) mission. It is anticipated that, to capture bidirectional context, each position would learn to make use of contextual data from all positions.

*3.4. XLM-RoBERTa*

The multilingual version of RoBERTa is called XLM-RoBERTa [49], which was released by Facebook as an update to their XLM-100 model. It was trained on 100 languages from 2.5 TB of filtered common crawl data. The "RoBERTa" part in XLM-RoBERTa originates from the fact that it uses the identical training procedures as the monolingual RoBERTa model, with the Masked Language Model training objective. There is no ALBERT-style sentence order prediction or BERT-style Next Sentence prediction in XLM-RoBERTa.

## 3.5. Support Vector Machine

In the 1990s, Vapnik [50] proposed the Support Vector Machine (SVM). The supervised learning algorithm has been shown to be a reliable and well-performing model for classification tasks [26,51]. An SVM constructs a hyperplane between separated marginal lines of the nearest support vectors (input vector) of the classes. In the space, an optimal separating hyperplane is determined by maximizing the marginal distance. An SVM is a linear classifier that also performs a non-linear classification problem using the kernel function. In the non-linear classification problem, the algorithm maps the data into input vector and uses the kernel function to transform low dimensional input space to a higher dimension to solve non-linear problems. The maximum margin hyperplane is used for optimal classification of a new instance (class). The higher the marginal distance between the classes, the more generalizable the result is. Figure 3 below provides a visual representation of an SVM.

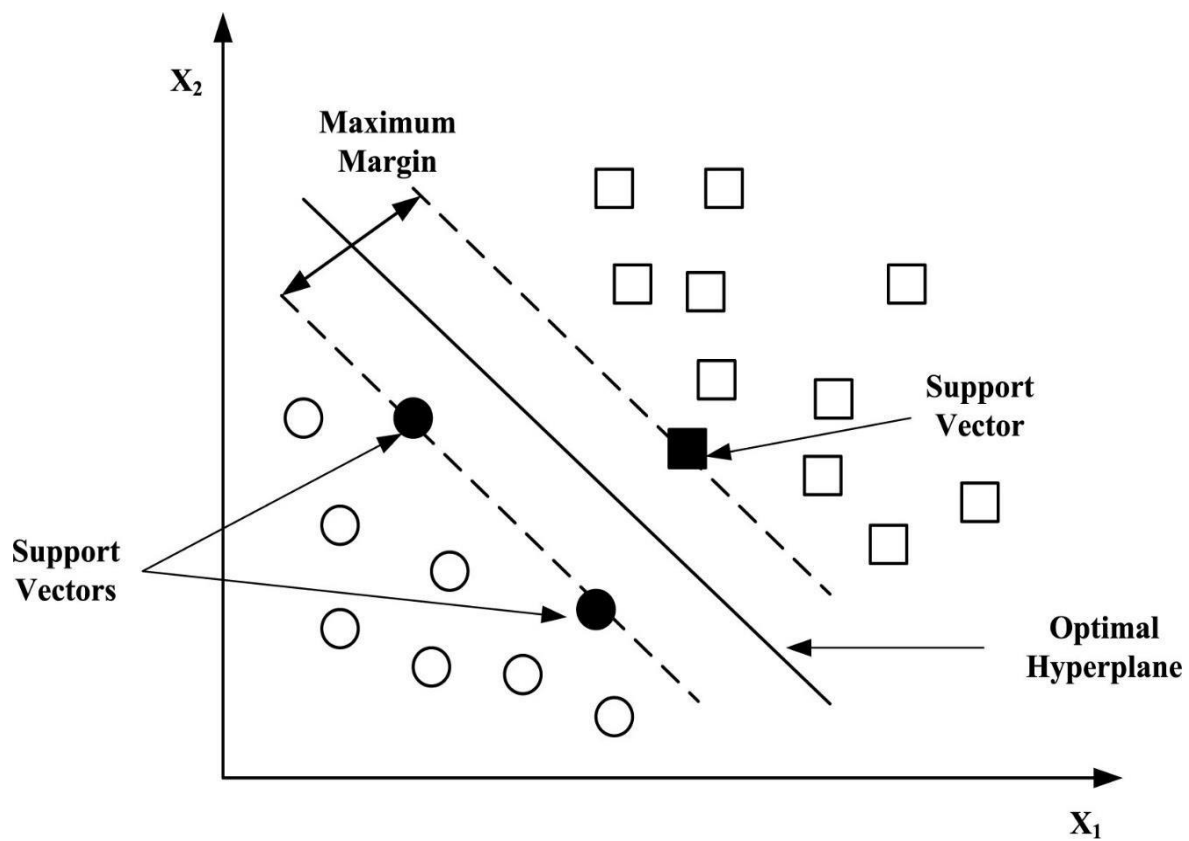

**Figure 3.** An illustration of a Support Vector Machine [51].

## 3.6. Random Forest

Breiman [52] introduced random forest (RF). RF is a robust machine learning algorithm that uses an ensemble learning approach. RF is an ensemble method that uses a random subset of features (from a training set) to train multiple independent decision trees (bootstrap) and predict instance by majority voting of each tree outcome or the average. The model is easy to interpret, runs efficiently on large database, is fast to train and scalable, performs well in complex datasets, and is robust to irrelevant features [53–55]. However, it is sensitive to overfitting, which can then be regulated using the numbers of trees. RF can be used for both classification and regression problems. Furthermore, the model can accommodate both categorical and continuous data and, most importantly, handles missing values and outliers well. Figure 4 below shows an illustration of the algorithm. The algorithm's iteration can be summarized as follows:

- Select k data points randomly;
- Build decision trees with the k data points;
- Predict y value based on aggregate of class (outcome);
- Voting or average.

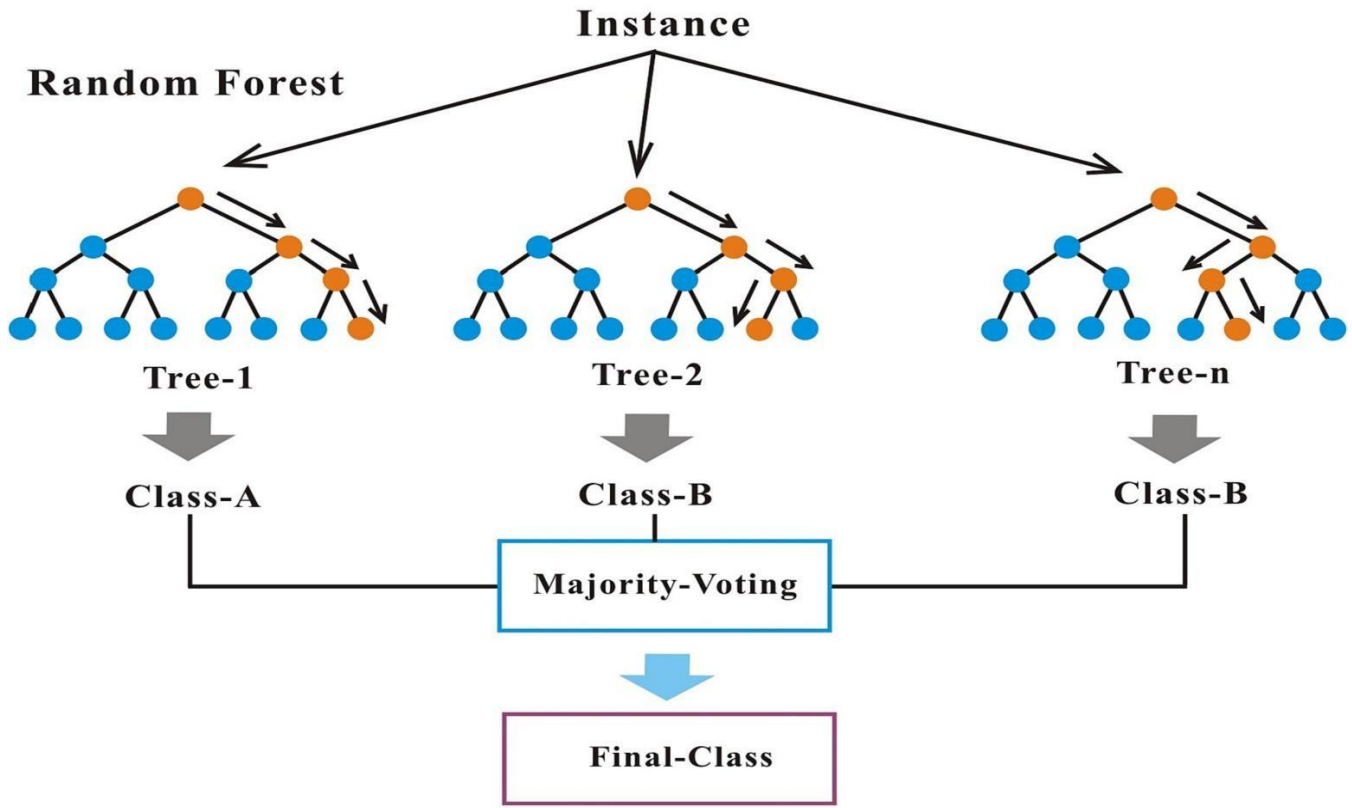

**Figure 4.** An illustration of random forest classifier [53].

*3.7. Evaluation Metrics*

This section discusses the evaluation metrics of the models to understand how well they have performed in this context and help decide on the best model. In a classification task, the common evaluation metrics are accuracy, precision, recall, and F1 measure. The proportion of correctly and wrongly labelled classes will be described in the form of true positive, true negative, false positive, and false negative.

Where:

True positive (TP): positive class classified correctly;

True negative (TN): negative class classified correctly;

False positive (FP): negative class wrongly predicted as positive (Type I error);

False negative (FN): positive class wrongly predicted as negative (Type II error);

Accuracy: the ratio of the number of samples predicted correctly to the total number of samples:

$$\frac{TP + TN}{TP + TN + FP + FN} \quad (1)$$

Precision: denoted as P is the percentage % of selected items that is correct.

$$\frac{TP}{TP + FP} \quad (2)$$

Recall: denoted as R is the percentage % of correct items that are selected.

$$\frac{TP}{TP + FN} \quad (3)$$

The F1 measure provides the balance between the precision and recall and can be denoted as

$$\frac{2PR}{P + R} \quad (4)$$

*3.8. Dataset*

In this study, we have used data from existing cyberbullying (CB) studies. The datasets were collected from Formspring.me and Twitter. We have named the datasets D1 and D2, for easy identification. In Dataset D1, we used the dataset of Agrawal and Awekar [28]. They collected the data from Formspring and employed three human annotators from the Amazon Mechanical Turk service to label the data as bullying or non-bullying. The data is publicly available via this link: https://github.com/sweta20/Detecting-Cyberbullying-Across-SMPs (accessed 24 April 2023). Table 1 below shows the class distribution of the dataset (D1).

**Table 1.** D1—Formspring (CB) data distribution [28].

| Label | Count |
| --- | --- |
| Not Cyberbullying (CB) | 11,997 |
| Cyberbullying | 776 |

Wang et al. [56] prepared a new dataset from six existing studies [10,28,32,57–59] Their datasets were manually annotated into fine-grained CB target classes, namely, victim's age, ethnicity, gender, religion, other quality of CB, and not cyberbullying (notcb). The datasets were imbalanced; hence, they applied a modified Dynamic Query Expansion (DQE) to augment the dataset in a semi-supervised manner. They then randomly sampled approximately 8000 tweets from each class to have a balanced dataset of approximately 48,000 tweets. The data set is publicly available, and can be accessed via the link below. Table 2 below provides the class distribution of their dataset https://drive.google.com/drive/folders/1oB2fan6GVGG83Eog66Ad4wK2ZoOjwu3F (accessed 24 April 2023).

**Table 2.** Twitter (CB) data distribution [56].

| Label | Count |
| --- | --- |
| Age | 7992 |
| Ethnicity | 7961 |
| Gender | 7973 |
| Religion | 7998 |
| Other CB | 7823 |
| Not CB | 7945 |

The issue of data scarcity and class imbalance is a popular problem in the CB detection domain. To resolve this, we took a different approach to Wang et al. [56]. This is because we aim to prepare a dataset that is comparable to prior and future work, and to enhance the development of CB detection algorithms. Thus, we prepared our binary classification dataset D2 from existing CB studies, including Wang et al. [56]. In dataset D2, we converted the multi-class dataset of Wang et al. [56] into binary classes by labelling the 'age', 'ethnicity', 'gender', 'religion', and 'other cyberbullying' classes as "bullying". Agrawal and Awekar [28] have 11,997 CB non-bullying instances, and only 776 bullying instances from FormSpring. Thus, we concatenated the two imbalanced binary class datasets to create ours. The instances that had less than three words or more than 100 words have been considered outliers and removed to obtain a more representational dataset. Table 3 below presents the distribution of our dataset.

**Table 3.** D2—distribution (Our study).

| Label | Count | Source | Annotation |
|---|---|---|---|
| Bullying | 19,553 | FormSpring + Twitter | Manual |
| Non-Bullying | 19,526 | FormSpring + Twitter | Manual |

### 3.9. Experimental Setup

The term frequency inverse document frequency (TF-IDF) feature vectors and Sentence BERT (SBERT) embeddings [60] are used as input to the traditional machine learning models. The TF-IDF was implemented using both unigram and bigram (ngram_range = 1, 2), the min_df is 5, and the max_df set to 0.8. The SBERT model used was 'all-MiniLM-L6-v2´. We deployed random forest and support vector machine algorithms using scikit-learn package. The latter used 'radial basis function' (RBF) as the kernel function with C = 5 and gamma = 0.5. The random forest classifier was employed with the number of estimators as 100, criterion is 'gini', and the bootstrap parameter set to 'True' (to allow taking subsamples for each decision tree). We employed GridSearchCV package for the parameter tuning for optimal solution. Additionally, we utilized HuggingFace transformers to implement pre-trained language models. The number of training epochs used is four. We have provided links to our source code (in the statement section) to evidence details of the transformer architecture implemented. For training and testing, we divided the data into stratified samples of 90% and 10%. With all the above descriptions for the models, the study was conducted in two cases. In case 1, we used dataset D1, which is highly imbalanced with 10,702 negative class instances and 703 positive class instances for training. This imbalanced dataset was given as input to the pre-trained language models, namely BERT, RoBERTa, XLNET, and XLM-RoBERTa. Similarly, in case 2, the dataset D2 was used as input to the ML algorithms, and we present the result in the subsequent section.

### 4. Experimental Results

This section presents the results of our experimental comparison of the classification algorithms. Our experiment was twofold. In the first experiment, we implemented the algorithms with the imbalanced dataset (D1), and named this case 1. In the second experiment, we implemented the algorithms with the balanced dataset (D2) prepared, and this is named case 2. Table 4 below presents the evaluation report of our case 1.

**Table 4.** Evaluation report of classification algorithms (case 1).

| Algorithms | Class | Training Size | Accuracy (%) | Precision (%) | Recall (%) | Macro F1-Score (%) |
|---|---|---|---|---|---|---|
| TF-IDF + RF | 0 | 10,792 | 0.80 | 0.89 | 0.75 | 0.82 |
| | 1 | 703 | | 0.41 | 0.27 | 0.34 |
| TF-IDF + SVM | 0 | 10,792 | 0.84 | 0.87 | 0.85 | 0.86 |
| | 1 | 703 | | 0.49 | 0.43 | 0.46 |
| BERT | 0 | 10,792 | 0.96 | 0.98 | 0.98 | 0.98 |
| | 1 | 703 | | 0.66 | 0.63 | 0.64 |
| XLNet | 0 | 10,792 | 0.95 | 0.97 | 0.99 | 0.98 |
| | 1 | 703 | | 0.70 | 0.41 | 0.52 |
| RoBERTa | 0 | 10,792 | 0.95 | 0.98 | 0.98 | 0.98 |
| | 1 | 703 | | **0.63** | **0.68** | **0.66** |
| XLM - RoBERTa | 0 | 10,792 | 0.94 | 0.98 | 0.96 | 0.97 |
| | 1 | 703 | | 0.53 | 0.68 | 0.60 |

In our experiment, the positive class instances ('bully') are denoted as '1´, and this class is of interest. Firstly, we implemented the traditional machine learning classifiers, namely support vector machine (SVM) and random forest (RF). The results in Table 1 above shows that RoBERTa achieved the best performance, with an F1 score of 0.66. Other LLMs achieved comparable results, except XLNet.

The traditional machine learning models performed poorly. We performed hyper-parameter tuning of the models for optimal performance. Unfortunately, there was no significant improvement. The optimal models achieved varied F1 scores at a range of ±0.064 to the published results in Table 1. In general, the algorithms struggled with the positive class instances ('bully') when compared to negative instances ('non bully'), especially the XLNet model. This is unsurprising, as its due to the class imbalance of the dataset (D1). However, RoBERTa showed a better result. The performance might be due to the improvement of RoBERTa on language understanding as the model is trained on larger text, bigger batch sizes, and longer sequence compared to BERT. Our results are superior to the results in previous studies. Agrawal and Awekar [28] applied bidirectional long short-term memory (BiLSTM) with an attention mechanism, and their implementation achieved an F1 score of 0.51 for the positive class instance ('bully'). In comparison to our study, all our large language models (LLMs) applied to D1 showed better performances.

Furthermore, in the experiments of Agrawal and Awekar [28], they improved on D1 by oversampling the minority class using the Synthetic Minority Oversampling technique (SMOTE). Their BiLSTM implementation achieved an F1 score of 0.91 for the positive class instance ('bully'). However, Emmery et al. [6] criticised their implementation. Emmery et al. [6] reproduced their experiment and discovered the overlap between train–test data due to the oversampling method. They showed the results of Agrawal and Awekar [28] are not reliable for the oversampled case. Thus, Emmery et al. [6] modified the implementation (BiLSTM with an attention mechanism) by oversampling only the training data, and achieved F1 score of 0.33 for the 'bully' class. To conclude for case 1 (D1), the LLMs, namely BERT, RoBERTa, XLNet, and XLM-RoBERTa, achieved better performance than that of Agrawal and Awekar [28] and Emmery et al. [6], as reported in Table 4 above. The performance of the LLMs can be attributed to their power to understand context better and effective understanding of long sequence text. Thus, it is not surprising to see that the models have performed better than the traditional machine learning models and the hybrid algorithms.

Table 5 below presents case 2 of our experiment. Using our dataset (D2), we ascertain RoBERTa as the state-of-the-art model, as the algorithm achieved the best performance overall, with an F1 score of 0.87 for the positive class ('bully'). Also, the results of all four language models prove that they are all comparable in performance with not much variance. This agrees with the result of Paul and Saha [44] that showed Bidirectional Encoder Representation from Transformer (BERT) performs better than deep learning algorithms like BiLSTM for automated cyberbullying detection. Our experimental findings showed that pre-trained language models are powerful and competitive with other models in the detection of cyberbullying on social media sites. Furthermore, it is worth noting that because we have used a balanced training dataset (D2), the traditional machine learning models also showed good performance; most notably, the support vector machine (SVM) achieved an F1 score of 0.85 for the positive class ('bully'). This is consistent with the results of Ogunleye [26] which showed SVM is a robust classification algorithm when fed with a balanced training dataset. In summary, we propose the use of RoBERTa for CB detection due to its consistent performance in both experiments (cases 1 and 2).

Table 5. Evaluation report of classification algorithms (case 2).

| Algorithms | Class | Training Size | Accuracy (%) | Precision (%) | Recall (%) | Macro F1-Score (%) |
|---|---|---|---|---|---|---|
| BERT | 0 | 17,573 | 0.85 | 0.85 | 0.86 | 0.86 |
|  | 1 | 17,598 |  | 0.86 | 0.85 | 0.86 |
| XLNet | 0 | 17,573 | 0.86 | 0.88 | 0.84 | 0.86 |
|  | 1 | 17,598 |  | 0.84 | 0.88 | 0.86 |
| RoBERTa | 0 | 17,573 | **0.87** | 0.87 | 0.86 | 0.86 |
|  | 1 | 17,598 |  | **0.86** | **0.87** | **0.87** |
| XLM-RoBERTa | 0 | 17,573 | 0.86 | 0.86 | 0.86 | 0.86 |
|  | 1 | 17,598 |  | 0.86 | 0.86 | 0.86 |
| SBERT + SVM | 0 | 17,573 | 0.85 | 0.84 | 0.87 | 0.86 |
|  | 1 | 17,598 |  | 0.87 | 0.83 | 0.85 |
| SBERT + RF | 0 | 17,573 | 0.81 | 0.79 | 0.86 | 0.82 |
|  | 1 | 17,598 |  | 0.84 | 0.77 | 0.81 |
| TF-IDF + SVM | 0 | 17,573 | 0.84 | 0.84 | 0.86 | 0.85 |
|  | 1 | 17,598 |  | 0.86 | 0.83 | 0.85 |
| TF-IDF + RF | 0 | 17,573 | 0.84 | 0.80 | 0.90 | 0.85 |
|  | 1 | 17,598 |  | 0.88 | 0.78 | 0.83 |

## 5. Conclusions

In this study, we aimed to ascertain a state-of-the-art (SOTA) language model for automated cyberbullying (CB) detection. We prepared a new dataset (D2) from existing CB studies. The datasets were originated from FormSpring and Twitter, and were manually annotated. We used the dataset in our implementation, and our results showed RoBERTa performed well in both experiments, cases 1 and 2. For a classification task, we argue that large language models (LLMs) can predict the minority class better than the approach using the traditional machine learning approach and/or the oversampling technique (case 1). This is due to the ability of the language model to understand the context of long and short text. In addition, we showed that RoBERTa perform better than deep learning approaches like BiLSTM with an attention mechanism. We also evidenced that, when the dataset is balanced, the traditional machine learning approach produces good performance; however, RoBERTa yielded a state-of-the-art (SOTA) performance. To conclude, our contributions can be summarised as follows. We prepared a new dataset (D2) for the development of algorithms in the field of CB detection. The dataset (D2) has been made publicly available for research access and use. We demonstrated how large language models can be used for automated CB detection with two datasets (D1 and D2). We presented SOTA results for CB detection by fine tuning RoBERTa.

In theory, the use of machine learning algorithms yields poor performance when fed with imbalanced datasets compared to large language models. Similarly, language models yield better results with balanced datasets. This implies that the performance of RoBERTa is consistent across different categorises (balanced or not) of cyberbullying datasets. In practice, our application is useful for social network owners, the government, and developers to implement cyberbullying detection algorithms to prevent and reduce the act. It is worth mentioning that our experiment is limited to English text. For future work, our implementation can be extended to other languages; most especially, the LLMs can be tuned with external non-English corpus for improving the SOTA models in non-English contexts. In addition, we consider implementing a multimodal approach to develop and enhance algorithms for CB detection. Furthermore, the implementation can be adopted to detect other forms of online abuse, including hate speech and cyber molestation.

**Author Contributions:** Conceptualization, B.O. and B.D.; methodology, B.O. and B.D.; software, B.D.; validation, B.O. and B.D.; formal analysis, B.D.; investigation, B.O. and B.D.; resources B.D.; data curation, B.O. and B.D.; writing—original draft preparation, B.O. and B.D.; writing—review and editing, B.O.; visualization, B.O. and B.D.; supervision, B.O.; project administration, B.O.; All authors have read and agreed to the published version of the manuscript.

**Funding:** This research received no external funding.

**Institutional Review Board Statement:** Not applicable.

**Informed Consent Statement:** Not applicable.

**Data Availability Statement:** The dataset and code used in this work is available on the Github repository (please see the link below) GitHub—Babitha23/Cyberbullying-detection Case 1: https://github.com/Babitha23/Cyberbullying-detection/tree/main/Case1. Case 2: https://github.com/Babitha23/Cyberbullying-detection/tree/main/Case2.

**Conflicts of Interest:** The authors declare no conflict of interest.